\def\year{2020}\relax
\documentclass[letterpaper]{article} 
\usepackage{subfigure}
\usepackage{algorithm}
\usepackage{amsmath,amsthm}
\usepackage{threeparttable}
\usepackage{booktabs,multirow}
\usepackage[noend]{algorithmic}

\usepackage{aaai20}  
\usepackage{times}  
\usepackage{helvet} 
\usepackage{courier}  
\usepackage[hyphens]{url}  
\usepackage{graphicx} 
\usepackage{graphicx}  
\newcommand{\nop}[1]{}
\newtheorem{definition}{Definition}[section]

\newtheorem{problem statement}{\bf Problem Statement}
\newtheorem{example}{Example}
\newtheorem{runningExample}{Running Example}

\title{Towards Combinational Relation Linking over Knowledge Graphs}
\author{%
{{Weiguo Zheng${^1}$}, Mei Zhang{${^2}$}}%
\vspace{1.6mm}\\
\fontsize{10}{10}\selectfont\itshape $~^{1}$Fudan University, China;\\ \fontsize{10}{10}\selectfont\itshape $~^{2}$Wuhan University of Science and Technology, China\\
\vspace{0.05in}
 \fontsize{10}{10}\selectfont\ttfamily\upshape \hspace{-0.2in}  zhengweiguo@fudan.edu.cn, zhangmeiontoweb@gmail.com
\\}

\begin{document}

\maketitle

\begin{abstract}
%
Given a natural language phrase, relation linking aims to find a relation (predicate or property) from the underlying knowledge graph to match the phrase. It is very useful in many applications, such as natural language question answering, personalized recommendation and text summarization. However, the previous relation linking algorithms usually produce a single relation for the input phrase and pay little attention to a more general and challenging problem, i.e., combinational relation linking that extracts a subgraph pattern to match the compound phrase (e.g. mother-in-law). In this paper, we focus on the task of combinational relation linking over knowledge graphs. To resolve the problem, we design a systematic method based on the data-driven relation assembly technique, which is performed under the guidance of meta patterns. We also introduce external knowledge to enhance the system understanding ability. Finally, we conduct extensive experiments over the real knowledge graph to study the performance of the proposed method.

\end{abstract}

\section{Introduction}\label{sec:introduction}



Knowledge graphs have been important repositories to materialize a huge amount of structured information in the form of triples, where a triple consists of $\langle$subject, predicate, object$\rangle$ or $\langle$subject, property, value$\rangle$. There have been many such knowledge graphs, e.g., DBpedia \cite{DBLP:conf/semweb/AuerBKLCI07}, Yago \cite{DBLP:conf/www/SuchanekKW07}, and Freebase \cite{DBLP:conf/sigmod/BollackerEPST08}. In order to bridge the gap between unstructured text (including text documents and natural language questions) and structured knowledge, an important and interesting task is conducting relation linking over the knowledge graph, i.e., finding the specific predicates/properties from the knowledge graph that match the phrases detected in the sentence (also may be a question).

Relation linking can power many downstream applications. As a friendly and intuitive approach to exploring knowledge graphs, using natural language questions to query the knowledge graph has attracted a lot of attentions in both academia and industrial communities \cite{DBLP:conf/emnlp/BerantCFL13,DBLP:conf/coling/BaoDYZZ16,DBLP:conf/acl/DasZRM17,DBLP:journals/tkde/Hu0YWZ18,DBLP:conf/wsdm/HuangZLL19}. %
Generally, the simple questions, e.g., who is the founder of Microsoft, are easy to answer since it is straightforward to choose the predicate ``founder'' from the knowledge graph that matches the phrase ``founder'' in the input question.
However, many questions are difficult to deal with due to the intrinsic variability and ambiguity of natural language.

\begin{figure}[t]
\begin{center}
    \includegraphics[scale=0.55]{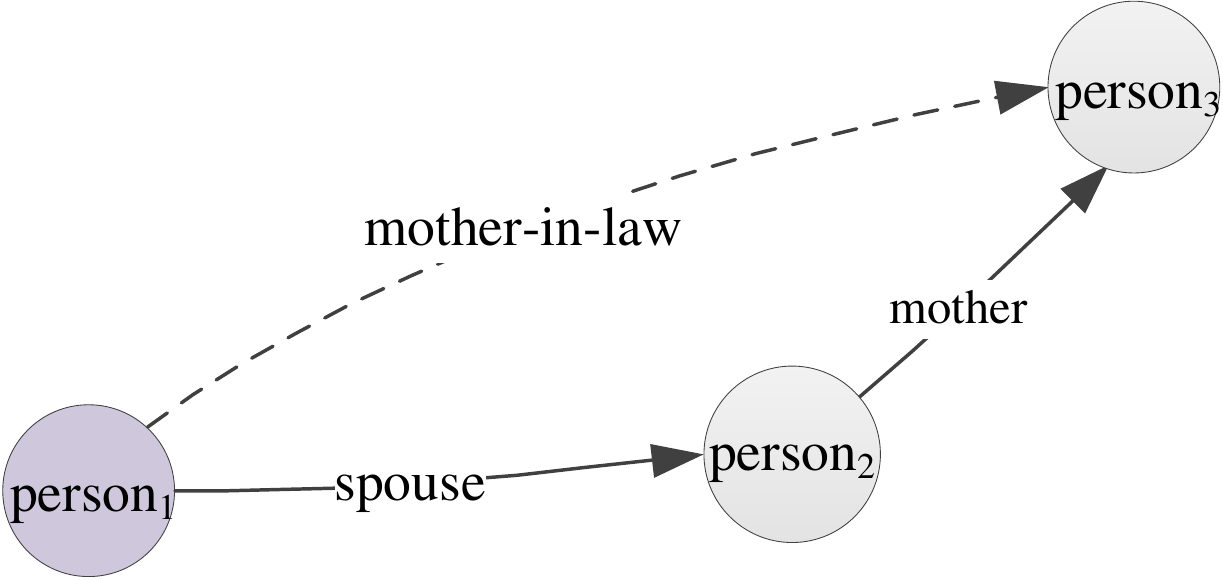}
   \caption{\small Example of combinational relations matching the compound phrase mother-in-law.}
   \label{fig:A running example}
   \vspace{-0.15in}
\end{center}
\end{figure}

\begin{runningExample}\label{runningExample1}
Let us consider the question ``Who is the mother-in-law of Barack Obama?''. It may be hard to answer when there is no predicate/property that directly matches the phrase ``mother-in-law''. 
Acutually, the combinational predicates ``mother'' and ``spouse'' should be inferred as matches. Precisely, it can be represented as the mother of one's spouse as depicted in Figure~\ref{fig:A running example}, where the dash line does not exist in the underlying knowledge graph.
\end{runningExample}

For ease of presentation, we do not distinguish predicates and properties in the following discussion unless it is necessary. Besides natural language question answering, relation linking can be helpful to many other applications such as personalized recommendation \cite{DBLP:conf/recsys/CatherineMEC17} and text summarization \cite{inproceedingsICALIP}.

Intuitively, finding the mapping predicates for input phrases can be considered a similarity search problem. Specifically, delivering the results by computing the similarity (or some distance measurements) between phrases and candidate predicates. Traditionally, the Levenshtein distance is used to measure the difference between two strings \cite{DBLP:journals/Levenshtein}. However, a predicate should link to the phrase although their surface form distance is large. For instance, the predicate ``spouse'' matches the phrase ``married to'', ``wife of'' or ``husband of'', but their Levenshtein distance is large. Moreover, it fails to distinguish
two literally similar strings but describe different semantic meanings, e.g., attitude and latitude. In order to overcome the two problems above, word embedding models are widely used to improve the relation linking performance. There are also some works resorting to external taxonomies like Wordnet\footnote{https://wordnet.princeton.edu/}. The synonyms, hyponyms, and variations are extracted to enhance the matching from phrases to predicates in the knowledge graph \cite{DBLP:conf/clef/BeaumontGL15,DBLP:conf/i-semantics/MulangSO17}. Another stream of researches perform entity linking (identifying the entity from the target knowledge graph that matches the input phrase) and relation linking as a joint task rather than taking them as separate tasks \cite{DBLP:conf/acl/YihCHG15,DBLP:conf/semweb/DubeyBCL18,inproceedingsISWC19}.

However, the existing relation linking systems aim to extract one predicate to match the input phrase, which may decrease the overall performance when multiple predicates are required to match a single phrase. For instance, as shown in the Running Example~\ref{runningExample1}, the phrase ``mother-in-law'' matches a path in the knowledge graph. For simplicity, a phrase $p$ is called a \textit{compound phrase} if it can be grounded to a group of predicates or properties which as a whole match the phrase $p$. To enhance the ability of understanding compound phrases in the view of knowledge graphs, we study the problem of combinational relation linking in this paper. Specifically, \emph{finding a subgraph pattern from the knowledge graph to match the input phrase}.
Notice that we just focus on the relation linking task and do not take entities into consideration as the entities may be unavailable in the input text or query. For example, a natural language question or keyword query is not required to contain entities.



\noindent \textbf{\emph{Challenges and Contributions}.} Actually, the traditional relation linking is a special case of our proposed combinational relation linking since only one edge pattern (i.e., the predicate/property) is detected. Nevertheless, the algorithms designed for traditional relation linking cannot be used to solve the combinational relation linking directly.
%
In order to preform combinational relation linking, it is required to address the following two challenges.

\noindent \emph{Challenge One: The gap between the phrase and the desired subgraph pattern.} Different from the single edge pattern, the desired mapping for the compound phrase is a subgraph pattern. In contrast, the input phrase consist of a sequence of words or even a single word, e.g., the phrase ``grandfather'' may match the subgraph pattern ``$\overrightarrow{person_1, father, person_2, father, person_3}$''. Thus we need to devise an effective mechanism to bridge the gap.

\noindent \emph{Challenge Two: It is difficult to determine how many predicates/properties in a match.} The target is to infer a subgraph pattern for the input compound phrase, but it is unknown that what the matched pattern is and how many edges (predicates and properties) the pattern contains, which increases the difficulty of conducting combinational relation linking.

Let us consider the process of manually performing relation linking. When the expert does not understand the input compound phrase, she/he may resort to some dictionary or search engine to make it clear. Inspired by the process of manual relation linking, we propose to use external knowledge to bridge the representation gap between phrases and subgraph patterns. The external knowledge, e.g., Oxford Dictionary API\footnote{https://developer.oxforddictionaries.com/} or Wikipedia\footnote{https://www.wikipedia.org/}, is invoked to interpret the phrase $p$ when $p$ is not understood by the system. Even if the phrase can be better understood by employing the side information, it remains a challenging problem to ground the phrase to a subgraph pattern since the structure is oblivious.  In order to determine the subgraph pattern, we design a group of meta patterns based on which the target subgraph pattern can be retrieved in a recursive manner.

In summary, we make the following contributions in this paper:
\begin{itemize}
    \item We design a systematic method to resolve the problem of combinational relation linking over knowledge graphs;
    \item We propose to use external knowledge to facilitate combinational relation linking;
    \item A recursive relation assembly technique based on meta patterns is devised to enhance the linking;
    \item Experimental results on two benchmarks show that our approach outperforms state-of-the-art algorihtms.
\end{itemize}


The rest of this paper is organized as follows. Section~\ref{sec:background} introduces the problem definition and framework of the approach.  Section~\ref{sec:Meta Pattern Recognition} introduces how to integrate external knowledge and defines several meta patterns. Section~\ref{sec:Compound Phrase Linking} presents the process of recursive relation assembly based on meta patterns.
The experimental results are provided in Section~\ref{sec:experiments}, followed by a brief review of related work in Section~\ref{sec:related work}. Finally, Section~\ref{sec:conclusion} concludes the paper.


\section{Problem Definition and Framework}\label{sec:background}

\begin{figure*}[t]
\begin{center}
    \includegraphics[scale=0.45]{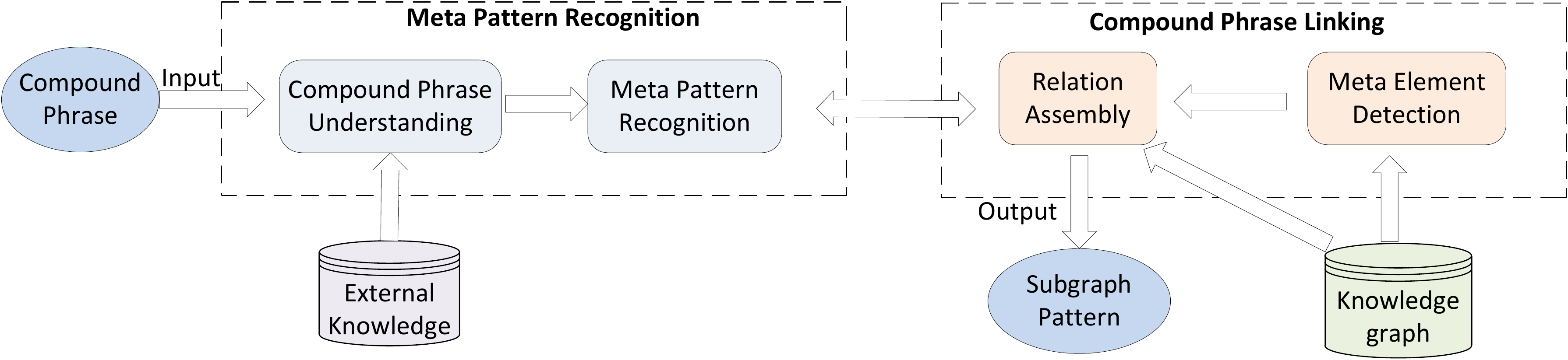}
   \caption{\small Framework of the approach.}
   \label{fig:framework}
   \vspace{-0.1in}
\end{center}
\end{figure*}


\subsection{Problem Definition}


In this section, we first review some basic notions and then give the framework of the approach. In the paper, the {\emph{knowledge graph}} is defined as Definition~\ref{def:knowledge graph}. For instance, $\langle$\emph{Oswald Lange, birthPlace, Germany}$\rangle$ is a triple in DBpedia. There is a special predicate ``type'' for each entity whose object is a type (e.g., Actor or Movie).

\begin{definition}\label{def:knowledge graph}
(\textbf{\emph{Knowledge graph, denoted by $G$}}). A directed graph consisting of triples $\langle$\emph{subject, predicate, object}$\rangle$ or $\langle$\emph{subject, property, value}$\rangle$, where subjects/objects are entities, and predicates/properties are relations.
\end{definition}

\begin{definition}
  (\emph{\textbf{Subgraph pattern}}). A subgraph pattern corresponds to a subgraph of the knowledge graph $G$, where each node $v$ is labeled as its type if $v$ corresponds to an entity in $G$.
\end{definition}

Figure~\ref{fig:A running example} presents a subgraph pattern. Note that the node of subgraph pattern is not necessary to correspond to an entity in $G$. For example, the node can be a literal.

\begin{definition}\label{def:compound phrase}
(\emph{\textbf{Compound phrase}}). A phrase $p$ is called a compound phrase with regard to the knowledge graph $G$ if $p$ can match a subgraph pattern in $G$.
\end{definition}

As shown in Definition~\ref{def:compound phrase}, the compound phrase is a relative term in terms of the target knowledge graph $G$. For instance, the phrase ``mother-in-law'' is not compound if the underlying knowledge graph $G$ contains the corresponding predicate that directly describes this relation. Thus the task of the paper is defined as next.

\begin{problem statement}
  Given a compound phrase $p$ and the underlying knowledge graph $G$, extracting the combinational relations (including predicates and properties), i.e., a subgraph pattern, from $G$ to match the phrase $p$.
\end{problem statement}

\subsection{Framework}

The overview of the proposed approach is depicted in Figure~\ref{fig:framework}. It consists of two components, i.e., meta pattern recognition and compound phrase linking.

In the first component, we introduce a group of meta patterns that can power the relation linking. Given a compound phrase, we can obtain its concrete meanings through external knowledge, e.g., a sentence interpreting the phrase. Then the meta pattern of the interpretation sentence is recognized.

In the second component, the subgraph pattern is constructed by filling the slots (nodes and edges) in the meta pattern generated above. The construction proceeds in a recursive manner.

\section{Meta Pattern Recognition}\label{sec:Meta Pattern Recognition}


In this section, we first perform compound phrase understanding, and then define the meta patterns. Finally, we discuss how to recognize the meta pattern of an interpretation sentence for the input compound phrase.


\subsection{Compound phrase understanding}
Due to the gap between unstructured natural language and  knowledge graph $G$, the compound phrase may not directly map a subgraph in $G$. We adopt external knowledge, e.g., Wikipedia, OXford Dictionary API, and Cambridge Dictionary API\footnote{https://dictionary.cambridge.org/zhs/}, to explain these relation phrases into simple sentences which describe concrete meanings of the phrases. For instance, Wikipeida gives the explanation of ``mother-in-law'' as: ``A mother-in-law is the mother of a person's spouse''. It is clear that the explanation provides more information of the input phrase, which is helpful to the extraction of desired subgraph pattern.

\subsection{Meta Pattern}
In this subsection, we introduce some meta patterns which could facilitate the relation linking task. As discussed above, it is a challenging task to determine the structure of the desired subgraph pattern directly. To resolve the problem, we propose a recursive assembly mechanism to construct the match based on several meta patterns. In principle, the meta patterns are very limited and can be enumerated in advance.

\begin{definition}\label{def:meta pattern}
(\emph{\textbf{Meta pattern}}). A meta pattern consists of two edges at most, where all the nodes and edges are unlabeled.
\end{definition}

\begin{figure}[b]
\begin{center}
    \vspace{-0.05in}
    \includegraphics[scale=0.5]{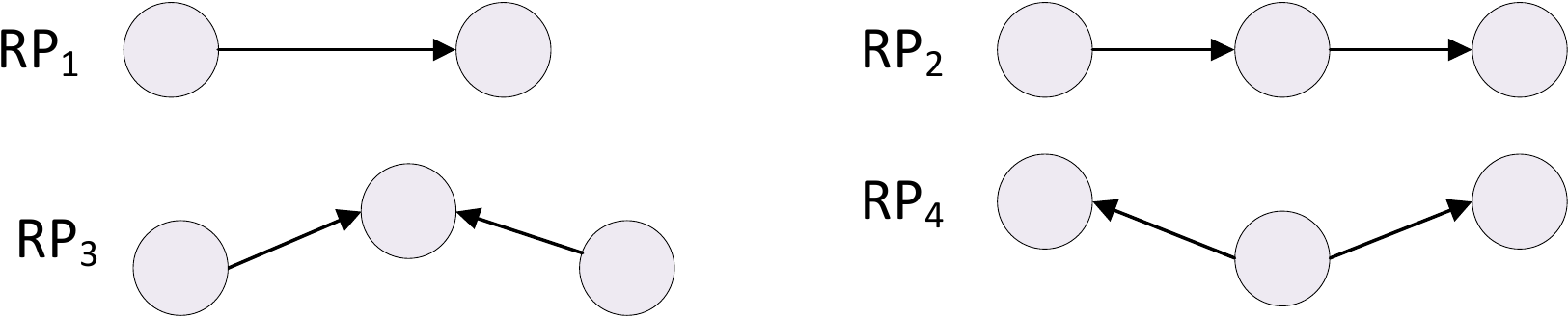}
   \caption{\small Meta patterns.}
   \label{fig:meta patterns}
\end{center}
\end{figure}

 %
There are four meta patterns as presented in Figure~\ref{fig:meta patterns}, where pattern $RP_2$ represents a progressive relationship, e.g. the pattern in Figure~\ref{fig:A running example} describes the compound phrase ``mother-in-law''; pattern $RP_3$ represents a converging coordinative relation, e.g. the phrase ``kinfolk'' (person from the same family); pattern $RP_4$ represents a diverging coordinative relation, .e.g. the phrase ``sportsman'' (a gender who plays sport).

Generally, each subgraph pattern can be assembled based on these meta patterns. Note that the pattern $RP_1$ corresponds to the traditional relation linking that deliver a single predicate or property in the knowledge graph. Actually, each subgraph pattern can be assembled through only pattern $RP_1$. However, it will increase the difficulty of inferring the structure for the interpretation sentence, which decreases the linking performance in further. On the other hand, larger meta patterns (e.g., a mete pattern consists of three or more edges) rarely occur in an explanation sentence directly. Moreover, larger meta patterns are difficult to recognize. Therefore, two-size meta patterns are good balance of representation ability and recognition difficulty.

\subsection{Meta Pattern Classification}\label{subsec:Meta Pattern Classification}


 Since there are only four meta patterns, recognizing the meta pattern of an explanation sentence can be taken as a classification problem. Since pattern $RP_1$ corresponds to a single predicate or property, it is can be identified through traditional relation linking algorithms. Hence, we just consider how to determine the other three meta patterns in the following discussion.

 As we know that there have been many classification models available, e.g., RNN Attention \cite{DBLP:conf/naacl/YangYDHSH16} and Text CNN \cite{DBLP:conf/emnlp/Kim14}, it is not the focus of this paper. One important issue is to collect training data, i.e., compound phrases, the corresponding explanation sentences, and the matched subgraph patterns. To best of our knowledge, there is no such training dataset yet. Since the knowledge graph may contain millions of triples, it is not a trivial task to manually build the training dataset. Therefore, we propose a data-driven approach to collecting training examples at a low cost.

\noindent \textbf{Training data collection - a data-driven approach.}
First of all, we need to address the challenge that how to conceive a compound phrase. Actually, titles of Wikipedia webpages provide a huge number of phrases, such as ``brother'', ``family man'', and ``hometown''. However, there may be noisy data as the titles may correspond to entities (e.g., Ludwig van Beethoven)  or even cannot be matched in the knowledge graph (e.g., Dominican Order). Thus these phrases can be directly discarded. Algorithm~\ref{alg:Training Example Collection} presents the details of collecting training examples. The external dictionary API is invoked to provide an explanation of the phrase. If there are just two relations $r_1$ and $r_2$ extracted from $G$ to match the simple phrases in $sen$, we need to check whether they are directly connected in $G$. In order to avoid the ambiguous matches, the corresponding subgraph pattern is added into $TE$ when $r_1$ and $r_2$ can form only one pattern. The relation linking of simple phrases will be discussed in the next section. The procedure proceeds until the number of identified training examples exceeds a threshold $\kappa$.

After obtaining the automatically generated training examples, it is easy to refine these results. Actually, the method of constructing training data above can be roughly used to perform combinational relation linking. It is also compared as a baseline in our experiments.

 \begin{algorithm}[tb]
\caption{Data-driven Relation Linking}
\label{alg:Training Example Collection}
\textbf{Input}: Wikipedia titles, dictionary API and knowledge graph $G$\\
\textbf{Output}: Training examples $TE$\\
\vspace{-0.18in}
\begin{algorithmic}[1] 
\STATE $TE \leftarrow \emptyset$
\FOR{each Wikipedia title $p$}

        \IF{$|TE| > \kappa$}
             \RETURN $TE$
        \ENDIF

        \IF{$p$ matches a/an property/predicate/type/entity in $G$}
            \STATE continue
        \ENDIF

        \STATE $sen \leftarrow$ the explanation sentence of $p$ by invoking dictionary API
        \STATE $R \leftarrow$ the relations corresponding to simple phrases in $sen$

        \IF{$|R| =2 $}
            \IF{$r_1$ and $r_2$ are adjacent in $G$ and form only one pattern}
                 \STATE  $TE \leftarrow$  subgraph pattern consisting of $r_1$ and $r_2$
            \ENDIF
        \ENDIF
\ENDFOR

\RETURN $TE$

\end{algorithmic}
\end{algorithm}

\noindent \textbf{Meta pattern classification.}
The state-of-the-art model RNN Attention model \cite{DBLP:conf/naacl/YangYDHSH16} is adopted in the experiments. Note that the aim is to infer the meta pattern of the sentence with regard to a specific knowledge graph. Generally, it may produce distinct meta patterns even for the identical sentence when the underlying knowledge graphs are different. In order to make the explanation sentence fit the model better, we include some features that depend on the underlying knowledge graph, i.e., perform relation mask over the sentence. For ease of presentation, if a phrase directly matches a single predicate or property, it is called a \emph{simple phrase}. Given an explanation sentence, we identify all the simple phrases and replace them with their matching predicates in the knowledge graph. Since the original
predicate IRIs may be too long and contains some noisy notations, we remove the prefix of each IRI and use a special symbol ``*'' to take the place of prefix. At the same time, the special symbol denotes the beginning of a relation (predicate or property). In other words, the input of the model is a variegated text.

 \begin{example}
   Let us consider the explanation sentence ``a male child''. We can obtain ``a foaf\footnote{http://xmlns.com/foaf/0.1/}:gender dbo\footnote{http://dbpedia.org/ontology/}:child'' through relation linking. Replacing the prefix with ``*'' leads to the sentence ``a *gender *child''. Feeding it to the classification model, we can obtain the progressive pattern $RP_2$.
 \end{example}


\section{Compound Phrase Linking}\label{sec:Compound Phrase Linking}



In this section, we first present the techniques of meta pattern filling (Section~\ref{subsec:Meta Pattern Filling}) and then conduct relation assembly according to the recognized meta patterns (Section~\ref{subsection:Relation Assembly}).

\subsection{Meta Element Detection}\label{subsec:Meta Pattern Filling}

Since the desired subgraph pattern consists of types and relations (predicates or properties),
we identify the meta elements including both node labels (i.e., types) and edge labels (i.e., relations) in this subsection.


\noindent (1) \emph{\textbf{Type Restriction}}. Type restriction step is to recognize type mention from the explanation sentence and link them to the knowledge graph $G$. The identified type will be the restriction of the pattern node. Actually, type is also relation. In this paper, we retrieve all type IRIs from the given knowledge graph and materialize them as a type dictionary. Each candidate phrase mention in the explanation sentence is enumerated to search over the type dictionary. 

\noindent (2) \emph{\textbf{Simple Phrase Linking}}. This step is to extract relation mentions from simple relation sentences and map these relation mentions to the knowledge graph $G$. These relations correspond to the edges of meta patterns. There are a variety of resources and systems for single relation linking. SIBKB \cite{singh2017capturing} provides searching mechanisms for linking natural language relations to knowledge graphs. BOA \cite{gerber2011bootstrapping} can be used to extract natural language representations of predicates independent of the language if provided with a Named Entity Recognition service. ReMatch \cite{DBLP:conf/i-semantics/MulangSO17} is an independently reusable tool for matching natural language relations to knowledge graph properties. EARL \cite{DBLP:conf/semweb/DubeyBCL18} is a recent approach for joint entity and relation linking which treats entity and relation linking as a single step. It determines the best semantic connection between all keywords of the question by exploiting the connection density between entity and relation candidates. All of these tools can be used to conduct single relation linking. In this paper, we ground the relation mentions extracted from the phrase explanation sentence to the predicates/properties through based on SIBKB.

\begin{example}
  Let us consider the compound phrase ``mother-in-law'' and its explanation ``the mother of a person's spouse''. We can extract type keyword ``person'' and link it to the type ``dbo:Person'' in the knowledge graph (DBpedia). It acts the type restriction of a node in the meta pattern. Conducting single relation linking, we can extract relation mentions ``mother'' and ``spouse'' from this sentence, and link them to ``dbo:mother'' and ``dbo:spouse'', respectively.  They correspond to edges in the meta patterns.
\end{example}


\subsection{Relation Assembly}\label{subsection:Relation Assembly}

 \begin{algorithm}[tb]
\caption{CompoundPhraseLinking($S$)}
\label{alg:Compound Phrase Linking}
\textbf{Input}: Phrase explanation sentence $S$\\
\textbf{Output}: Subgraph pattern $sp$\\
\vspace{-0.15in}
\begin{algorithmic}[1] 
    \STATE $T, R \leftarrow$ meta elements in $S$

     \IF{$S$ contains compound phrase $p$}
            \STATE $S' \gets$ explanation sentence of $p$
            \STATE $sp' \gets$ CompoundPhraseLinking$(S')$ \\
            \STATE $S \gets$ Replace $p$ with $sp'$ in $S$\\
     \ELSIF{$|R| = 1$}
        \RETURN $R$
     \ELSIF{$|R| > 1$}
        \STATE $MP \gets$ meta pattern of $S$\;
        \STATE    $SG \gets$ Data-driven and meta pattern checking\\
        \RETURN $SG$\\
     \ENDIF
     \RETURN ``No match''


\end{algorithmic}
\end{algorithm}

%
%

With the recognized meta elements and meta pattern, we are ready to produce the subgraph pattern. A naive approach is performing relation assembly following a data-driven paradigm. Specifically, the subgraph pattern is constructed by retrieving the subgraphs that cover all the meta elements, which is similar to the procedure in Algorithm~\ref{alg:Training Example Collection}. However, it may produce ambiguous subgraphs, i.e., distinct subgraphs cover the meta elements and follow the meta pattern. Thus the overall linking performance will degrade correspondingly.

\noindent \textbf{Meta Pattern Assembly. }
In order to address the problem above, we propose a novel approach to assembling relations under the guidance of meta patterns and the underlying knowledge graph. Algorithm~\ref{alg:Compound Phrase Linking} depicts the process. Unlike the ways mentioned above, we add meta patterns to restrict subgraphs which can improve the precision of data-driven method. Given the phrase explanation sentence $S$, we infer its meta pattern through the classification model as shown in Section~\ref{sec:Meta Pattern Recognition}. Based on the meta pattern, we assemble the recognized meta elements (including types and relations) as a subgraph one by one in the order they appear in the explanation sentence. If the assembled subgraph pattern matches a subgraph in the knowledge graph, it will be delivered as the result; Otherwise, we will modify the order of the recognized relations and perform the similar checking above. 

\begin{example}
Let us consider the running example. By applying the pattern classification model, we can infer the pattern of explanation sentence ``the mother of a person's spouse'' is the progression pattern $RP_2$ as presented in Figure~\ref{fig:meta patterns}. Then we can assemble relations ``dbo:mother'' and ``dbo:spouse'' derived from meta element detection step according to the progression pattern. Thus there are two assembled subgraphs $\langle$x dbr:mother z$\rangle$, $\langle$z dbr:spouse y$\rangle$  and $\langle$x dbr:spouse z$\rangle$, $\langle$z dbr:mother y$\rangle$. Nevertheless, based on the meta pattern constraint, we can infer that the compound phrase ``mother-in-law'' can be represented as $\langle$person$_1$ dbr:spouse person$_2$$\rangle$, $\langle$person$_2$ dbr:mother person$_3$$\rangle$.
\end{example}

\noindent\textbf{Nested Pattern Assembly.} Some explanation sentences not only contain simple phrases, but also compound phrases. The compound phrase in the explanation sentence is called  a ``nested phrase''. As shown in Algorithm~\ref{alg:Compound Phrase Linking}, the nested compound phrase is parsed recursively.

\begin{example}
Let us consider the compound phrase ``great-grandparent''. Its explanation sentence is ``a parent of your grandparent'', which contains another compound phrase ``grandparent''. So we need to parse ``grandparent'' first. Based on meta pattern assembly, we can infer the subgraph pattern of ``grandparent'' is $\langle$person$_1$ dbo:parent person$_2$$\rangle$, $\langle$person$_2$ dbo:parent person$_3$$\rangle$. Then, ``grandparent'' is taken as a new simple relation ``dbo:grandparent''. Then we can identify the left modified explanation sentence with the classification model. It follows the progressive pattern as well. Finally, we can deliver the subgraph pattern $\langle$person$_0$ dbo:parent person$_1$$\rangle$, $\langle$person$_1$ dbo:parent person$_2$$\rangle$, $\langle$person$_2$ dbo:parent person$_3$$\rangle$ for the phrase ``great-grandparent''.
\end{example}


\section{Experimental Study}\label{sec:experiments}

The proposed approach is systematically studied in this section over real datasets. Section~\ref{subsec:datasets} presents the experimental settings, followed by the results in Section~\ref{subsec:results}.

\subsection{Experimental Settings}\label{subsec:datasets}

\noindent \textbf{Datasets}. To evaluate the performance of our approach, we collect some compound phrases based on Wikipedia webpage titles.  The details of collecting the training examples are described in Section~\ref{subsec:Meta Pattern Classification}. In the experiments, DBpedia is adopted as the knowledge graph. Finally, we collect 600 compound phrase, where 500 phrases are used to train the model and 100 phrases are used to test the performance. Beyond that, we also collect 100 simple phrases to evaluate the effect of the proposed external knowledge and meta patterns. All of these collected data will be released once the review process is complete.


 \noindent \textbf{Competitors}.
To evaluate the performance of proposed approach, we compare it with the following competitors.

\begin{itemize}
  \item Keyword Match: It just simply matches the compound phrases to all predicates in the knowledge graph. A predicate will be delivered once it matches the input phrase.
  \item SIBKB \cite{singh2017capturing} provides searching mechanisms for linking natural language relations to knowledge graphs.
  \item Similarity Search:  It calculates the similarity between the compound phrase and each predicate and then returns the best predicate with the highest similarity.
  \item Data-driven linking: Similar to our approach, it is equipped with external knowledge and exploits the data-driven approach to retrieve subgraph patterns. The only difference is that it works without the guidance of meta patterns.
\end{itemize}

\noindent \textbf{Evaluation metrics}.
We evaluate the effectiveness (precision, recall, and F1-measure) and efficiency (the response time from receiving a phrase to delivering its matches) of the methods.

\subsection{Experimental Results}\label{subsec:results}

\begin{table}[t]
\caption{Effect of relation mask on meta pattern classification}
\label{table1}
\begin{center}
\begin{tabular}{|l|c|c|c|c|}
	\hline Method &Precision & Recall& F-score\\
	\hline Without relation mask & 0.73 & 0.78& 0.72\\
    \hline With relation mask &\textbf{0.90} & \textbf{0.88} & \textbf{0.86}\\
    \hline
\end{tabular}
\end{center}
\end{table}

\begin{table}[t]
\caption{Results of competitors and our approach on compound phrase linking}
\label{table3}
\begin{center}
\begin{tabular}{|l|c|c|c|c|}
	\hline Method &Precision & Recall& F-score\\
	\hline Keyword Match & 0.050 & 0.025& 0.033\\
	\hline Similarity Search & 0.167 & 0.083& 0.094\\
    \hline SIBKB & 0.050 & 0.050& 0.048\\
	\hline Data-driven Linking& 0.167 & \textbf{0.808}& 0.150\\
    \hline \textbf{Our approach} & \textbf{0.65} & \textbf{0.625} & \textbf{0.633}\\
    \hline
\end{tabular}
\end{center}
\end{table}

\begin{table}[t]
\caption{Evaluation of external knowledge on simple phrase linking}
\label{table2}
\begin{center}
\begin{tabular}{|l|c|c|c|c|}
	\hline Method &Precision & Recall& F-score\\
	\hline Without Explanation & 0.20 & 0.175 & 0.183\\
    \hline \textbf{With Explanation} & \textbf{0.80} & \textbf{0.775} & \textbf{0.783}\\
    \hline
\end{tabular}
\end{center}
\end{table}

\noindent  \textbf{\emph{Evaluation of meta pattern classification}}. As discussed in Section~\ref{subsec:Meta Pattern Classification}, we include some features that depend on
the underlying knowledge graph, i.e., replacing phrase mentions with the corresponding relations. Table~\ref{table1} shows its effect of the performance of classifying meta patterns.
The precision of the meta pattern classification can achieve 0.90 when is equipped with relation mask. Moreover, the recall improves  as well.  It indicates performing relation mask that takes advantage of the target knowledge graph is very effective as the F-score achieves 0.16 gain.

\noindent  \textbf{\emph{Results of competitors and our approach}}.
Table~\ref{table3} shows the results of the four competitors and our approach. We can find that methods keyword match and similarity search performs very poorly with low precision and recall. That is because they can just link a phrase to a single predicate or property. However, most compound phrases match subgraph patterns with multiple edges rather than a single relation pattern directly. Another reason of SIBKB performing poorly is that it depends on PATTY database to find out synonyms for relation keywords. However, PATTY database contains a very limited number of synonyms. The performance will degrade greatly once it does not contain the compound phrases.
Though data-driven linking achieves a relatively high recall, its precision rather low, which the overall F-score correspondingly. In contrast, our approach powered by meta patterns performs much better than the data-driven linking as it achieves a good balance between precision and recall.


\noindent  \textbf{\emph{Evaluation of external knowledge on simple phrase linking}}.
In order to study the importance of external knowledge, i.e., obtaining the concrete meanings of the input compound phrase, we also evaluate its effect on simple phrase linking that extracts a single relation for an input simple phrase.
 As shown in Table~\ref{table2}, it is clear that the performance of relation linking enhanced with explanation outperforms that does not considering external knowledge significantly. Hence, exploiting external knowledge is helpful to both simple and compound phrase linking.

\noindent  \textbf{\emph{Response time of different methods}}. We also report the time cost of each method as presented in Table~\ref{table4}, where the response time is averaged over 100 testing compound phrases. We can see that keyword match runs the fastest as it only performs exact matching computation. Without the guidance of meta patterns, the data-driven linking faces a larger search space. Thus it consumes more time than the other methods, which can further illustrate the superiority of meta patterns.


\begin{table}[t]
\caption{Response time of different methods}
\label{table4}
\begin{center}
\begin{tabular}{|l|c|c|c|c|}
	\hline Method & Time cost (sec)\\
	\hline {Keyword Match} & \textbf{0.161} \\
	\hline Similarity Search & 1.661\\
	\hline Data-driven Linking & 3.065\\
    \hline {Our approach} & \textbf{0.377} \\
    \hline
\end{tabular}
\end{center}
\end{table}

\noindent \textbf{\emph{Error Analysis}}. 
In order to improve the compound phrase linking in the future, we analyze the results and categorize the errors into three groups, i.e., errors in
meta pattern classification, relation assembly and meta element identification.

Classification errors. As our proposed method highly depends on meta patterns, the final result will be incorrect when the predicted pattern is false. For example, let us consider the compound phrase ``countrywoman'' in the experiments. Its explanation sentence is ``a woman from your own country''. The predicted pattern by the classification model is the progressive pattern $RP_2$. Based on the meta pattern $RP_2$, we can obtain the assembled subgraph pattern \{$\langle$x dbo:country z$\rangle$, $\langle$z foaf:gender y$\rangle$\}. However, the correct meta pattern of the sentence should be the diverging coordinate pattern $RP_4$, and the desired subgraph pattern is \{$\langle$x dbo:country z$\rangle$, $\langle$x foaf:gender y$\rangle$\}.

Relation assembly errors. Relation assembly is not a trivial task especially for nested compound phrases.
 For example, the explanation sentence of the compound phrase ``co-sister'' is ``the wife of your husband's brother'' (denoted by $S_1$), where ``brother'' is also a compound phrase with regard to DBpedia. The explanation of ``brother'' is ``a male sibling'' (denoted by $S_2$). By performing relation assembly, the system delivers the subgraph pattern \{$\langle$x dbo:relative z$\rangle$, $\langle$z foaf:gender y$\rangle$ \} for ``brother''. Then ``dbo:brother'' is taken as a simple relation in sentence $S_1$. Finally, the system returns
the assembled subgraph pattern \{$\langle x$ dbo:spouse $y_1\rangle$, $\langle y_1$ dbo:relative $y_2 \rangle$, $\langle y_2$ foaf:gender $y_3 \rangle$, $\langle y_3$ dbo:spouse $y_4 \rangle$\}. Since dbo:relative and dbo:spouse have assembled with progression pattern, not foaf:gender and dbo:spouse, the result is incorrect.

Errors caused by meta element identification. There may be some noisy types and relations in the detected meta elements, which increases difficulties for relation assembly as it is hard to distinguish them from the correct ones. For example, the explanation of phrase ``stepmother'' is ``the woman who is married to someone's father but who is not their real mother'', where relation ``dbo:mother'' is extracted as a match of the mention ``mother''. Nevertheless, the relation dbo:mother should not be detected and used for the downstream relation assembly.

\section{Related Work}\label{sec:related work}

Performing relation linking is highly related to knowledge graph completion \cite{ebisu2019graph}. Hence, we give a brief review of algorithms for relation linking and knowledge graph completion next.


\subsection{Relation Linking}
The previous work on relation linking can be divided into two groups, i.e., independent relation linking \cite{nakashole2012patty,singh2017capturing,DBLP:conf/i-semantics/MulangSO17} and joint relation linking \cite{DBLP:conf/semweb/DubeyBCL18,inproceedingsISWC19,sakor2019old}.

Independent relation linking. PATTY \shortcite{nakashole2012patty} uses iterative bootstrapping strategies to extract RDF resources from unstructured text. However, PATTY cannot be used directly as a component for relation linking in a QA system and needs to be modified according to the application. SIBKB \cite{singh2017capturing} uses PATTY as the underlying
knowledge source and proposes a novel approach based on the semantic similarity between mentions and predicates/properities. ReMatch \cite{DBLP:conf/i-semantics/MulangSO17} employs dependency parse characteristics with adjustment rules and then carries out a match against knowledge graph properties enhanced with the lexicon Wordnet. However, the time efficiency is relatively low for each question.

Joint relation linking. A bunches of approaches address entity linking and relation linking jointly \cite{miwa2014modeling,wang2018joint,DBLP:conf/semweb/DubeyBCL18,inproceedingsISWC19,sakor2019old}.
EERL \cite{inproceedingsISWC19} computes relation candidates based on identified entities. Sakor et al. present an approach for jointly linking entities and relations within a short text into the entities and relations of DBpedia \cite{sakor2019old}. It uses the context of entities for finding relations and does not require training data. Miwa and Sasaki propose a history-based structured learning approach that jointly extracts entities and relations in a sentence \cite{miwa2014modeling}. EARL \cite{DBLP:conf/semweb/DubeyBCL18} determines the best semantic connection between all keywords of the question by exploiting the connection density between entity and relation candidates.

Most of the methods above can just perform simple phrase linking. Thus they exhibit poor performance to handle compound phrase linking that aims to extract a subgraph pattern from the knowledge graph.

\subsection{Knowledge Graph Completion}

Knowledge graph completion is proposed to predict the missing edges between any two entities in the knowledge graph. It is an alternative way to deal with compound phrase linking. A variety of algorithms performing knowledge graph completion have been proposed these years. The knowledge graph embedding based models, which embed entities and relations into a continuous space, are widely used, such as translation-based embedding models \cite{bordes2013translating,lin2015modeling} and manifold-based embedding model \cite{guo2015semantically,xiao2015one,ebisu2018toruse}. These methods suffer from the problem of result interpretation. Besides the embedding based approaches, there are a bunches of other methods, such as PRA models \cite{lao2011random,wang2016knowledge} and GPAR models \cite{fan2015association,ebisu2019graph}.



Most of the algorithms designed for knowledge graph completion cannot be directly used for handling compound phrase linking since
they can only predict relations that belong to a predefined relation dictionary. However, it is very likely that compound phrases correspond to out-of-vocabulary relations.

\section{Conclusion}\label{sec:conclusion}

In this paper, we study the problem of finding a subgraph pattern to match the given compound phrase, which has received little attention up to now is not a trivial task. To bridge the gap between unstructured natural language and enhance the system understanding ability, we introduce external knowledge in the linking process. As relation linking highly depends on the underlying knowledge graph, we propose a data-driven relation assembly technique. More importantly, we define several meta patterns which can guide the relation assembly. The systematic empirical results show that the proposed approach outperforms the competitors significantly. It also confirms the effectiveness of the introduction of external knowledge and meta patterns.



\bibliographystyle{aaai}
\bibliography{icde_sqp}

\end{document}